\documentclass[10pt,twocolumn,letterpaper]{article}

\usepackage[pagenumbers]{cvpr}
\usepackage{amsmath,amssymb,amsfonts}
\usepackage{amsthm}
\usepackage{pifont}
\usepackage{graphicx}
\usepackage{textcomp}
\usepackage{xcolor}
\usepackage{mathtools}
\usepackage{microtype}
\usepackage{balance}
\usepackage{ragged2e}
\usepackage{subcaption}
\usepackage{array}
\usepackage{booktabs}
\usepackage{tabularx}
\usepackage{tikz}
\usepackage{balance}
\usepackage{enumerate}
\usepackage{enumitem}
\usepackage{listings}
\usepackage{cvpr} 
\usepackage{hyperref}

\usepackage{algorithm}
\usepackage{algpseudocode}

\begin{document}

\title{Provenance-Driven Reliable Semantic Medical Image Vector Reconstruction via Lightweight Blockchain-Verified Latent Fingerprints}


\author{
\begin{tabular}{c c}
Mohsin Rasheed & Abdullah Al-Mamun \\
Augusta University & Augusta University \\
\texttt{mrasheed@augusta.edu} & \texttt{aalmamun@augusta.edu} \\
\end{tabular}
}

\maketitle


\begin{abstract}
Medical imaging is essential for clinical diagnosis, yet real-world data frequently suffers from corruption, noise, and potential tampering, challenging the reliability of AI-assisted interpretation. Conventional reconstruction techniques prioritize pixel-level recovery and may produce visually plausible outputs while compromising anatomical fidelity, an issue that can directly impact clinical outcomes. We propose a semantic-aware medical image reconstruction framework that integrates high-level latent embeddings with a hybrid U-Net architecture to preserve clinically relevant structures during restoration. To ensure trust and accountability, we incorporate a lightweight blockchain-based provenance layer using scale-free graph design, enabling verifiable recording of each reconstruction event without imposing significant overhead. Extensive evaluation across multiple datasets and corruption types demonstrates improved structural consistency, restoration accuracy, and provenance integrity compared with existing approaches. By uniting semantic-guided reconstruction with secure traceability, our solution advances dependable AI for medical imaging, enhancing both diagnostic confidence and regulatory compliance in healthcare environments.
\end{abstract}


\section{Introduction}

\subsection{Motivation}
Medical imaging plays a central role in modern diagnostics, with modalities such as MRI, CT, and X-ray enabling detailed visualization of internal anatomy~\cite{hussain2022modern}. The quality and integrity of these images directly affect clinical decisions, yet real-world imaging data are frequently degraded by noise~\cite{noise}, acquisition and compression artifacts~\cite{compression-1}, or even intentional tampering~\cite{tampering-1,tampering-2,tampering-3} and device-level attacks~\cite{yaqoob2019security}, all of which can undermine diagnostic reliability. Standard reconstruction methods~\cite{tampering-1,tampering-2,tampering-3,compression-1,dl-image-recovery} focus primarily on pixel-level recovery, which can restore visually plausible content but fail to preserve critical anatomical structures. This gap presents a significant challenge for dependable AI systems that aim to assist or automate diagnosis.

Recent advances in deep generative models, including U-Net~\cite{unet-1,unet-2}, GANs~\cite{gan-1,gan-2}, and diffusion-based architectures~\cite{defussion}, have improved inpainting and image restoration tasks. However, these methods remain largely agnostic to the underlying semantic content of medical images. In clinical scenarios, even small structural deviations can lead to incorrect interpretations~\cite{hussain2022modern}, making semantic fidelity as important as visual similarity. For example, a GAN‑based attack~\cite{attack-1} changed a lung‑cancer scan in seconds and escaped expert detection and a one‑pixel perturbation~\cite{attack-2} flipped diagnostic output. Furthermore, existing approaches~\cite{gan-attack-detect-1,gan-attack-detect-2,gan-attack-detect-3,attack-detect-4} lack mechanisms to verify or track the provenance of reconstructed images, which is crucial for auditing, reproducibility, and clinical trust.

\subsection{Proposed Approach and Challenges}
To address these limitations, we propose a semantic-aware reconstruction framework that operates directly on \textit{vector embeddings of medical images}, rather than solely on pixels, and leverages high-level latent fingerprints to guide restoration of corrupted data. By conditioning the model on these embeddings extracted from verified image data, our approach ensures that reconstructed structures remain consistent with clinically relevant patterns. In addition, we integrate a lightweight blockchain-based provenance layer, which anchors each reconstruction event in a verifiable ledger. Such integration establishes tamper-evident traceability, decentralized trust, and verifiable data lineage across reconstruction pipelines, essential properties~\cite{hasani2022trustworthy} for clinical imaging where authenticity and accountability are paramount. This combination allows not only accurate reconstruction but also traceable, auditable verification of every image modification, ensuring both reliability and trustworthiness.

Despite blockchain’s promise for ensuring trust and reliability, its direct adoption in medical imaging introduces several practical challenges. First, conventional blockchain consensus mechanisms~\cite{zhang2020cycledger, kokoris2018omniledger, al2022dean} such as PoW or BFT incur significant computational and latency overhead, incompatible with real-time or high-throughput image reconstruction tasks. Second, maintaining a complete ledger of reconstruction operations can impose excessive storage and bandwidth costs, especially when handling large-scale image datasets~\cite{wang2023lightweight}. Third, the high latency of block confirmation~\cite{nguyen2022latency, mamun2021baash} undermines the responsiveness needed for time-sensitive diagnostic scenarios. Finally, existing blockchain frameworks often assume persistent peer connectivity, whereas medical imaging workflows may span intermittently connected hospital systems or cloud-edge environments~\cite{alskaif2021blockchain, abdella2021architecture}. 

\begin{table*}[!t]
\centering
\caption{Comparison of state of the art systems vs \textsc{Prosima}.}
\label{tbl:comparison}
\resizebox{\textwidth}{!}{
\begin{tabular}{@{} l  c  c  c  l @{}}
\toprule
\textbf{Category} & \textbf{Semantic} & \textbf{Provenance} & \textbf{Sharding/Scale} & \textbf{Fidelity / Reliability} \\
\midrule

Vanilla U-Net 
& No & No & No 
& Strong low-level fidelity but semantically blind. \\

CNN/Unrolled Models
& Partial & No & No
& High fidelity but semantically blind. \\

Diffusion Models
& Indirect & No & No
& High realism but obscure traceability. \\

Semantic Vision–Language Encoders
& Yes & No & No
& Semantic alignment but no end-to-end reconstruction. \\

GAN-based Restoration
& Limited & No & No
& Realistic textures but untraceable. \\

Classical BFT Blockchains
& No & Yes (metadata only) & Partial
& No per-image verifiable reconstruction. \\

\midrule
\textbf{\textsc{Prosima}} (This work)
& \textbf{Yes} 
& \textbf{Yes} 
& \textbf{Yes} 
& \textbf{High semantic fidelity and full traceability.} \\
\bottomrule
\end{tabular}
}
\end{table*}

\subsection{Our Contributions}
To address these challenges, we design and implement provenance-secured semantic image reconstruction for medical applications (\textsc{Prosima}), a system that integrates a U-Net–based reconstruction backbone augmented with semantic consistency guidance from pre-trained embeddings (e.g., Contrastive Language–Image Pretraining-CLIP or medical feature extractors). By conditioning reconstruction on latent fingerprints anchored in a lightweight, scale-free blockchain, \textsc{Prosima} ensures tamper-evident verification, traceability, and clinically faithful reconstruction. The system introduces innovations in consensus efficiency, adaptive sharding, distributed load balancing, and storage optimization tailored for vector data. Our framework is designed for robustness and generalization across multiple datasets and corruption types. By explicitly incorporating semantic guidance, perceptual quality metrics, and blockchain-anchored fingerprints, the proposed method bridges the gap between conventional pixel-focused reconstruction and the stringent requirements of dependable AI in medical imaging. Experiments demonstrate superior performance in structural fidelity, semantic consistency, and provenance verification compared to state-of-the-art baselines.

In summary, the paper makes the following core technical contributions:

\textbf{(i) Semantic-aware Reconstruction Framework:} A semantic-aware, provenance-verified image reconstruction pipeline based on a hybrid U-Net backbone, augmented with high-level embeddings and latent fingerprints anchored on a lightweight blockchain, ensuring clinically faithful and tamper-evident reconstructions.

\textbf{(ii) Geographically Fragmented Ledger Consensus:} A set of lightweight blockchain protocols leveraging the \textit{Geographic Fragmentation Technique} (GFT) and the \textit{Scale-Free Graph}–based lightweight consensus mechanism to achieve efficient agreement and cost-effective distributed storage sharding for secure, auditable image management.

\textbf{(iii) System Prototype: } A fully implemented prototype evaluated on a large medical imaging dataset and diverse corruption scenarios, demonstrating superior structural fidelity, semantic consistency, and provenance verification compared to mainstream baselines, including: high PSNR ($\ge 28.2$~dB), strong SSIM ($\ge 0.86$), robust semantic alignment with embedding cosine similarity ($\ge 0.92$), a $100\%$ provenance verification rate, and low end-to-end verification latency of approximately $204$~ms per image even under severe Gaussian corruption (noise level $\sigma$ up to $0.10$).

\section{Related Work}
\subsection{Deep Learning for Medical Image Reconstruction}
Early deep networks for inverse problems treated reconstruction as post-processing of analytic or iterative solutions. FBPConvNet demonstrated strong artifact suppression for CT by learning a residual U-Net on top of filtered back-projection~\cite{jin2017fbpconvnet}. For MR, model-based unrolled methods became dominant: ADMM-Net unfolded an optimization scheme for CS-MRI and learned its proximal operators end-to-end~\cite{yang2016admmnet}. Variational Networks learned trainable regularizers and data-consistency steps for multi-coil MRI, achieving clinical-quality reconstructions~\cite{hammernik2018varnet}. Learned Primal–Dual further integrated the forward operator within a primal–dual unrolling, improving PSNR/SSIM over TV, U-Net denoisers, and FBP on tomographic tasks~\cite{adler2018lpd}. These lines establish that coupling physics with learned priors outperforms pure post-processing. Surveys such as McCann et~al.\ review the design patterns behind these CNN-based inverse solvers~\cite{mccann2017review}. 
PnP-ADMM replaced proximal steps with powerful denoisers, enabling modular priors with convergence analysis under bounded denoiser assumptions~\cite{venkatakrishnan2013pnp,chan2016pnp}. RED formalized denoiser-defined regularization, further broadening algorithmic choices~\cite{romano2017red}. These frameworks underpin many practical MRI/CT pipelines by flexibly injecting semantic or learned priors into reconstruction.

\subsection{Diffusion/Score-Based Priors for Inverse Problems}
Score-based generative models (a.k.a.\ diffusion models) have recently been used as powerful priors for reconstruction without task-specific supervision. Song et~al.\ trained a score model on medical images and solved CT/MRI inverse problems by posterior sampling guided by physics, showing competitive quality and stronger generalization across measurement models~\cite{song2021scoremed}. Subsequent works systematize diffusion for inverse problems and introduce operator-aware or blind formulations (e.g., diffusion posterior sampling, parallel priors for both image and operator) with strong results on MRI and sparse/limited-angle CT~\cite{dds_github, chung2023parallel}. These approaches motivate our semantic conditioning: diffusion priors can be steered by external cues (e.g., embeddings) to improve structure preservation.

\subsection{Semantic Guidance in Medical Imaging}
Semantic encoders provide higher-level anatomical constraints that surpass the limitations of pixel-level losses, enabling reconstruction models to preserve clinically relevant structures. In radiology vision–language (VL) pretraining, BioViL learns joint image–text representations from chest X-rays and corresponding reports, improving semantic alignment and downstream interpretability tasks~\cite{boecking2022biovil,bannur2023biovilt}. MedCLIP extends this idea by performing contrastive learning on unpaired medical images and reports, achieving strong zero-shot pathology recognition and robust visual–semantic representations for classification and retrieval~\cite{medclip_emnlp,medclip_pmc}. \textit{While these models enhance semantic awareness, they lack mechanisms for ensuring structural fidelity, provenance, or tamper-evident reconstruction}. 


%

\subsection{Provenance, Trustworthy AI, and Blockchain in Imaging}
Trusted medical imaging requires verifiable provenance throughout acquisition, reconstruction, and post-processing stages. Recent efforts toward trustworthy AI emphasize auditability, explainability, and data lineage as essential prerequisites for clinical deployment~\cite{hasani2022trustworthy}. In radiology, the ESR white paper highlights blockchain’s potential to ensure integrity and tamper-evidence along imaging data flows, particularly in clinical trials and regulatory pipelines~\cite{esr_blockchain_whitepaper}.

\subsection{Limitations of Current Solutions}
Table~\ref{tbl:comparison} highlights the shortcomings of current approaches. While deep learning–based reconstruction methods, including U-Net, diffusion priors, and semantic encoders, have achieved remarkable gains in structural preservation, \textit{they fundamentally operate as isolated learning systems with no guarantees on data integrity, reproducibility, or auditability}. U-Net, in particular, remains the most robust and interpretable backbone for medical image reconstruction~\cite{ronneberger2015u, jin2017fbpconvnet, du2020medical, azad2024medical}, effectively capturing both global context and fine anatomical detail through its encoder–decoder structure with skip connections, and consistently outperforming transformer- or GAN-based approaches across modalities such as CT~\cite{jin2017fbpconvnet} and MRI~\cite{hammernik2018varnet}. Compared to diffusion-based priors~\cite{song2021solving}, U-Net models achieve superior performance with lower computational cost; however, despite these strengths, \textit{no current U-Net–based approach explicitly ensures reliability, reproducibility, or provenance-secured workflows, leaving critical gaps for trustworthy clinical deployment}.

Although blockchain has emerged as a promising direction for securing medical data provenance, there is yet no framework that addresses the unique demands of reliable image reconstruction. Existing blockchain-based imaging systems primarily focus on metadata registration or access control rather than coupling provenance directly with the reconstruction process itself. Notwithstanding these advances, current blockchain frameworks still rely on heavyweight Byzantine Fault Tolerant (BFT) consensus mechanisms~\cite{castro1999pbft}. 
Although designs such as Tendermint~\cite{buchman2016tendermint} and HotStuff~\cite{yin2019hotstuff} improve responsiveness and linearity, they remain unsuitable for high-throughput, computationally intensive imaging workflows for several reasons: \textbf{(i)} these protocols introduce significant communication overhead and synchronization latency~\cite{mamun2021baash}, \textbf{(ii)} lack data-local adaptive sharding for large volumetric data, and \textbf{(iii)} are unable to guarantee real-time reliability and semantic fidelity during distributed reconstruction. \textit{These limitations collectively motivate the need for a lightweight, domain-optimized blockchain protocol that integrates directly with semantic-aware reconstruction to ensure tamper-evident, provenance-secured, and clinically faithful imaging.}




\begin{figure}[!t]
     \centering
    \includegraphics[width=85mm]{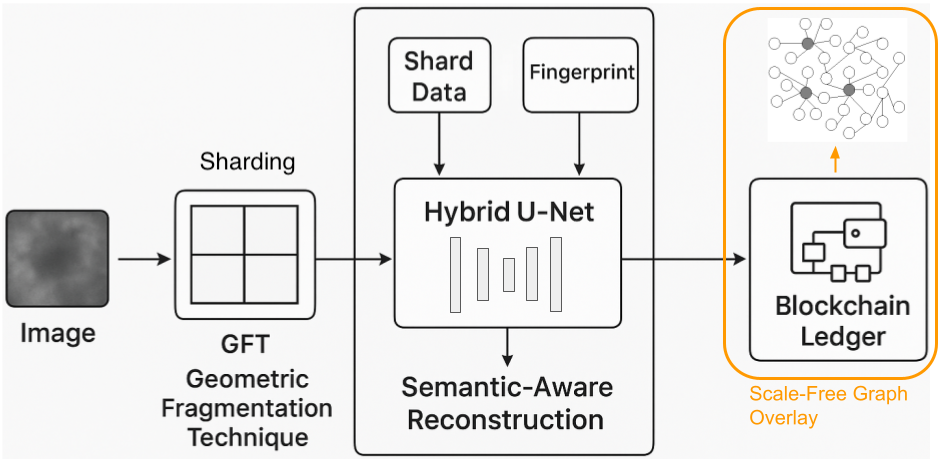}
    \caption{\textcolor{black}{Proposed \textsc{Prosima} framework for semantic-aware image reconstruction powered by lightweight blockchain ledger equipped with scale-free graph mechanism and geographic fragmentation technique.
    }}
      
    \label{fig:prosima-workflow}
     \vspace{-7mm}
\end{figure}

\section{Methodology}
\subsection{\textsc{Prosima} System Overview}
\textsc{Prosima} performs semantic-aware medical image reconstruction in a distributed environment, targeting three interdependent design goals: \textbf{(i)} computational reliability, \textbf{(ii)} semantic reconstruction quality, and \textbf{(iii)} scalability under high-throughput conditions. Figure~\ref{fig:prosima-workflow} illustrates the \textsc{Prosima} architecture that integrates three tightly coupled components: \textbf{(i)} a Geometric Fragmentation Technique (GFT) for deterministic sharding, a \textbf{(ii)} Hybrid U-Net for semantic reconstruction, and a \textbf{(iii)} scale-free blockchain layer for provenance and reliability.

Each medical image (2D or 3D) is partitioned into $K$ geometrically coherent shards using GFT, which enforces \textit{deterministic fragmentation boundaries}. Unlike random or content-based sharding, GFT guarantees reproducible shard topology across nodes, allowing the system to minimize redundant blockchain replication while ensuring fast shard localization and reaggregation. Deterministic shard IDs eliminate the need for global state synchronization, reducing storage duplication overhead from $O(N)$ to near $O(\log N)$ in the scale-free overlay.
These shards are distributed across nodes organized by a scale-free overlay graph~\cite{barabasi2003scale} with a power-law degree distribution, where high-degree leader nodes maintain both full ledgers and denser shard sets, while low-degree peers store sparse fragments. This configuration inherently supports fault-tolerant redundancy and minimizes synchronization latency.

Reconstruction occurs primarily at an edge node that retrieves only the required shards (locality-first) through a blockchain-guided lookup. Cached semantic embeddings at the edge accelerate reassembly by providing latent priors from previously verified reconstructions. The reconstruction engine employs a \textit{Hybrid U-Net}, which combines convolutional encoders for local texture recovery and transformer-based decoders for global semantic coherence. Conditioning on latent fingerprints retrieved from the blockchain ensures that reconstructed outputs are verifiable and semantically aligned with their original images. Finally, a compact Merkle-root fingerprint of each reconstructed image is anchored on-chain, enabling immutable provenance tracking with minimal blockchain overhead.

The integration of the scale-free blockchain layer addresses the shortcomings of conventional BFT-based systems (e.g., PBFT, Tendermint, HotStuff), which suffer from heavy communication and synchronization costs unsuitable for image-intensive workloads. Instead, \textsc{Prosima} employs a lightweight consensus mechanism over the scale-free network, reducing agreement complexity to sub-logarithmic levels and ensuring low-latency validation for provenance updates. This synergy between deterministic geometric fragmentation, hybrid semantic reconstruction, and blockchain-backed reliability forms the core of \textsc{Prosima}’s scalable and trustworthy medical imaging pipeline.


\subsection{Problem Definition}

Let $I_c \in \mathbb{R}^{H \times W}$ denote a corrupted medical image, and $I_o$ its corresponding original (ground-truth) image. In \textsc{Prosima}, each image is first partitioned into $K $ rectangular shards using the GFT:
\vspace{-2mm}
\begin{equation}
\{I_c^{(k)}\}_{k=1}^{K} = \text{GFT}(I_c),
\end{equation}

where each shard $I_c^{(k)}$ is distributed across nodes in a scale-free overlay graph. This fragmentation enables efficient storage, locality-aware retrieval, and parallel reconstruction across distributed nodes.

The goal of semantic-aware reconstruction is to learn a mapping:
\vspace{-2mm}
\begin{equation}
f_\theta: \{I_c^{(k)}\}_{k=1}^{K} \mapsto I_r,
\end{equation}

where $I_r$ is the reconstructed image that preserves both low-level pixel fidelity and high-level semantic structures. Reconstruction is conditioned on a latent fingerprint $z^*$ extracted from verified shards via the blockchain, enabling traceable provenance and reliable reconstruction. Formally, the task is posed as the following optimization problem:
\vspace{-1mm}
\begin{equation}
\begin{split}
\min_\theta \mathcal{L}_{total}(I_r, I_o) &= \mathcal{L}_{pixel}(I_r, I_o) + \lambda_1 \mathcal{L}_{perceptual}(I_r, I_o) \\
&\quad + \lambda_2 \mathcal{L}_{semantic}(I_r, I_o),
\end{split}
\end{equation}

where

\begin{itemize}
    \item $\mathcal{L}_{pixel}$ is a pixel-wise reconstruction loss (L1 or L2), enforcing low-level fidelity between $I_r$ and $I_o$, critical for medical details;
    \item $\mathcal{L}_{perceptual}$ is a perceptual loss computed over high-level features extracted from a pre-trained network (e.g., VGG-16), ensuring structural and textural similarity;
    \item $\mathcal{L}_{semantic}$ enforces semantic consistency via embeddings from a pre-trained encoder (e.g., Contrastive Language–Image Pretraining-CLIP or medical feature extractor), preserving anatomically meaningful structures.
\end{itemize}

In \textsc{Prosima}, reconstruction uses a \textit{hybrid U-Net generator} $G$ that is conditioned on both:

\begin{itemize}
    \item \mbox{$\{I_c^{(k)}\}$} retrieved from distributed nodes (locality-first)
    \item \mbox{$z^*$} anchored on the blockchain, encoding semantic priors and verified image characteristics.
\end{itemize}

The reconstructed image is then obtained as:

\begin{equation}
I_r = G\big(\{I_c^{(k)}\}_{k=1}^{K}, z^*\big),
\end{equation}

ensuring that $I_r$ is \textit{visually realistic, semantically faithful, and verifiable} via blockchain-based provenance.
This formulation explicitly unifies distributed storage, GFT-based sharding, hybrid U-Net reconstruction, and blockchain-verifiable latent fingerprints, providing a rigorous foundation for clinical reliability, semantic accuracy, and system scalability.

\begin{algorithm}[!t]
\floatname{algorithm}{Protocol}
\scriptsize
\caption{Federated Training of Semantic Reconstruction Models with Ledger Anchoring}
\label{alg:training_func}
\begin{algorithmic}[1]
\State \textbf{Input:} Training dataset $\mathcal{D}$ across edge nodes, encoder $E$, generator $G$, semantic extractor $S$, hash function $H$, blockchain ledger $L$
\Function{TrainAndAnchorModels}{$\mathcal{D}, E, G, S, H, L$}
    \For{each local node $n_i$ in parallel}
        \For{each mini-batch $(I_o, I_c)$ from $\mathcal{D}_i$}
            \State $I_r \gets G(I_c, S(I_c), E(I_o))$ \Comment{Generate reconstruction}
            \State Compute losses: 
            \[
            \mathcal{L}_{total} = \mathcal{L}_{pixel} + \lambda_1 \mathcal{L}_{perceptual} + \lambda_2 \mathcal{L}_{semantic}
            \]
            \State Update $(E,G,S)$ via Adam optimizer
        \EndFor
        \State $h_i \gets H(\text{model\_weights}_i)$ \Comment{Compute fingerprint}
        \State $tx_i \gets \text{CreateTransaction}(h_i, \text{metadata}_i)$
        \State \textbf{ConsensusProtocol}($tx_i$) \Comment{Anchor via ledger}
        \State $L.\text{commit}(tx_i)$
    \EndFor
    \State $(E,G,S) \gets \text{Aggregate}(\{(E_i,G_i,S_i)\})$
    \State \textbf{Output:} Trained models $E, G, S$ anchored in ledger
\EndFunction
\end{algorithmic}
\end{algorithm}

\begin{algorithm}[!t]
\floatname{algorithm}{Protocol}
\scriptsize
\caption{Latent Fingerprint Anchoring with Deterministic Fragmentation}
\label{alg:fingerprint_func}
\begin{algorithmic}[1]
\State \textbf{Input:} Image $I$, pretrained encoder $E$, hash $H$, ledger $L$, fragmentation operator $\mathcal{F}_{GFT}$
\State \textbf{Precondition:} $E$ obtained from Protocol~\ref{alg:training_func}
\Function{LatentFingerprint}{$I, E, H, L, \mathcal{F}_{GFT}$}
    \State $\{I_k\}_{k=1}^K \gets \mathcal{F}_{GFT}(I)$
    \For{each shard $I_k$}
        \State $z_k \gets E(I_k)$
        \State $h_k \gets H(z_k)$
        \State $tx_k \gets \text{CreateTransaction}(h_k, \text{metadata}_k)$
        \State \textbf{Consensus}($tx_k$)
        \State $L.\text{commit}(tx_k)$
    \EndFor
    \State $h_{root} \gets \text{MerkleRoot}(\{h_k\})$
    \State $L.\text{commit}(h_{root})$
    \State \textbf{Output:} Root fingerprint $h_{root}$
\EndFunction
\end{algorithmic}
\end{algorithm}

\subsection{Training Objective}

The reconstruction model is trained to jointly optimize multiple objectives ensuring pixel accuracy, perceptual fidelity, and semantic consistency. Formally, the total loss is defined as:
\vspace{-2mm}
\begin{equation}
\mathcal{L}_{total} = \mathcal{L}_{pixel} + \lambda_1 \mathcal{L}_{perceptual} + \lambda_2 \mathcal{L}_{semantic},
\end{equation}

where $\mathcal{L}_{pixel}$ enforces low-level fidelity (L1 or L2) between the reconstructed image $I_r$ and the original $I_o$, $\mathcal{L}_{perceptual}$ measures differences in high-level feature representations from a pre-trained network (e.g., VGG) to ensure structural and textural coherence, and $\mathcal{L}_{semantic}$ preserves anatomical structures and latent fingerprint consistency via embeddings from a task-specific or pre-trained encoder (e.g., CLIP or a medical feature extractor).
The model parameters are optimized using the Adam optimizer~\cite{kingma2014adam} with $\beta_1=0.9$, $\beta_2=0.999$, and a learning rate of $1\times10^{-4}$. Intermediate semantic embeddings or fingerprints may be cached locally on edge nodes to accelerate training. This formulation ensures that reconstructed images are clinically reliable, semantically faithful, and verifiable via blockchain-based provenance in a distributed environment.

\begin{algorithm}[!t]
\floatname{algorithm}{Protocol}
\scriptsize
\caption{Semantic-Aware Reconstruction Guided by Fragment Fingerprints}
\label{alg:semantic_func}
\begin{algorithmic}[1]
\State \textbf{Input:} Corrupted image $I_c$, pretrained models $E,G,S$, ledger $L$
\State \textbf{Precondition:} Models trained via Protocol~\ref{alg:training_func}, fingerprints anchored via Protocol~\ref{alg:fingerprint_func}
\Function{Reconstruct}{$I_c, E, G, S, L$}
    \State $\{I_{c,k}\}_{k=1}^K \gets \mathcal{F}_{GFT}(I_c)$
    \For{each shard $I_{c,k}$}
        \State $s_k \gets S(I_{c,k})$
        \State $z_k^* \gets \text{RetrieveLatent}(L, H(s_k))$
        \State $I_{r,k} \gets G(I_{c,k}, s_k, z_k^*)$
    \EndFor
    \State $I_r \gets \text{Reaggregate}(\{I_{r,k}\}, \mathcal{F}_{GFT}^{-1})$
    \State \textbf{Consensus}($I_r$) \Comment{Optional verification}
    \State \textbf{Output:} Reconstructed image $I_r$
\EndFunction
\end{algorithmic}
\end{algorithm}

\begin{algorithm}[!t]
\floatname{algorithm}{Protocol}
\scriptsize
\caption{Lightweight Scale-Free Consensus for Shard Verification}
\label{alg:ledger_func}
\begin{algorithmic}[1]
\State \textbf{Input:} Transactions $\{tx_k\}$ with fingerprint hashes $\{h_k\}$, shard assignments via GFT
\State \textbf{MPI Configuration:} $P$ parallel processes, shard-to-process mapping $\mathcal{M}$

\Function{ParallelConsensus}{$\{tx_k\}$}
    \State \textbf{// Phase 1: Parallel Shard Pre-verification (MPI ranks execute concurrently)}
    \For{each MPI rank $r \in [0, P-1]$ \textbf{in parallel}}
        \State $\text{shard\_set}_r \gets \{tx_k \mid \mathcal{M}(tx_k) = r\}$ \Comment{GFT-based locality mapping}
        \For{each $tx_k \in \text{shard\_set}_r$}
            \State $sig_{r,k} \gets \text{Sign}(tx_k, sk_r)$ \Comment{Local signature generation}
            \State Cache $(tx_k, sig_{r,k})$ in local buffer
        \EndFor
    \EndFor
    
    \State $\text{all\_sigs} \gets \text{MPI\_Allgather}(\text{local signatures})$ \Comment{O(log P) latency}
    
    \State \textbf{// Phase 2: Degree-Weighted Aggregation (executed in parallel)}
    \For{each MPI rank $r$ \textbf{in parallel}}
        \For{each $tx_k \in \text{shard\_set}_r$}
            \State $\text{weight\_sum} \gets \sum_{i} \text{degree}(n_i) \cdot \mathbb{I}[\text{sig valid}(sig_{i,k})]$
            \If{$\text{weight\_sum} \geq \text{threshold}(2f+1)$}
                \State Mark $tx_k$ as \textit{verified}
            \EndIf
        \EndFor
    \EndFor
    
    \State \textbf{// Phase 3: Leader Assembly (single coordinator, minimal serialization)}
    \If{rank = 0} \Comment{Only leader node assembles block}
        \State $block \gets \text{AssembleBlock}(\{tx_k\}_{\text{verified}}, \text{all\_sigs})$
        \State Broadcast $\langle \text{COMMIT}, block \rangle$ \Comment{O(log P) via MPI\_Bcast}
    \EndIf
    \State \textbf{MPI\_Barrier}() \Comment{Synchronize before ledger write}
    \State Append $block$ to local shard-specific ledger
    
    \State \textbf{Output:} Verified block with parallel execution time $O(\frac{n}{P} + \log P)$
\EndFunction
\end{algorithmic}
\end{algorithm}
\section{\textsc{Prosima} Protocols}

\subsection{Federated Training with Fingerprint Anchoring}

Protocol \ref{alg:training_func} performs decentralized training of the encoder–generator–semantic extractor triplet while cryptographically linking every local update to the blockchain ledger. Each edge node optimizes reconstruction losses on its local data and anchors a compact fingerprint of the resulting weights as a ledger transaction. This process guarantees model provenance, reproducibility, and tamper-evident training history, allowing subsequent inference and reconstruction stages to reference only verified model states.

\begin{table*}[!t]
\centering
\scriptsize
\caption{Reconstruction and provenance metrics (average ± std). Higher is better for PSNR, SSIM, Cosine, Verification Rate; lower is better for Latency. PROSIMA (full) achieves the best tradeoff, highest semantic cosine (0.92) and SSIM (0.86) while keeping latency acceptable because GFT reduces retrieval/replication overhead compared to the w/o-GFT variant.}
\label{tab:main_results}
\begin{tabular}{lccccc}

\toprule
Method & PSNR (dB) & SSIM & Embedding Cosine & Verification (\%) & Latency (ms) \\
\midrule
U\text{-}Net (pixel)              & $\ge 26.0$ & $\ge 0.80$ & $\ge 0.62$ & --   & $\le 251$ \\
U\text{-}Net + perceptual         & $\ge 27.6$ & $\ge 0.83$ & $\ge 0.68$ & --   & $\le 238$ \\
U\text{-}Net + semantic           & $\ge 28.0$ & $\ge 0.85$ & $\ge 0.87$ & --   & $\le 233$ \\
\textbf{\textsc{PROSIMA} (w/o GFT)}         & $\mathbf{\ge 28.1}$ & $\mathbf{\ge 0.85}$ & $\mathbf{\ge 0.90}$ & $ \textbf{100} $  & $\mathbf{\le 219}$ \\
\textbf{\textsc{PROSIMA} (full)}           & $\mathbf{\ge 28.2}$ & $\mathbf{\ge 0.86}$ & $\mathbf{\ge 0.92}$ & \textbf{100} & $\mathbf{\le 204}$ \\
\bottomrule
\end{tabular}
\vspace{-5mm}
\end{table*}

\vspace{-2mm}
\subsection{Latent Fingerprint Extraction and Fragmentation}
\vspace{-2mm}
Protocol \ref{alg:fingerprint_func} encodes each medical image into deterministic latent fragments using the GFT fragmentation operator. The encoder outputs shard-level embeddings whose hashes are individually committed to the ledger, forming a Merkle-root fingerprint that uniquely identifies the complete image. By anchoring every shard hash through the consensus network, the protocol ensures immutable provenance, traceable fragment lineage, and deterministic re-verification of any reconstructed image.

\vspace{-2mm}
\subsection{Semantic-Aware Reconstruction with Shard Retrieval}
\vspace{-2mm}
Protocol \ref{alg:semantic_func} reconstructs corrupted or incomplete medical images using semantically guided retrieval of verified latent shards. For each corrupted fragment, the semantic extractor produces a high-level representation that queries the ledger for its most consistent latent counterpart. The generator then synthesizes localized reconstructions which are re-aggregated into the final image. Because every retrieved latent vector originates from an anchored fingerprint, this stage achieves trust-preserving reconstruction with both pixel-level fidelity and semantic consistency.

\vspace{-2mm}
\subsection{Ledger Verification and Scale-Free Consensus}
\vspace{-2mm}
Protocol~\ref{alg:ledger_func} introduces an MPI-accelerated, scale-free consensus that performs shard verification in parallel across distributed ranks. Transactions are partitioned among processes using GFT-based locality mapping, and each rank independently verifies and signs its assigned shards. Partial signatures are then exchanged using non-blocking collective primitives (e.g., \texttt{MPI\_Allgather}, \texttt{MPI\_Bcast}), aggregated via degree-weighted quorum reduction, and assembled into blocks by the shard leader. This hybrid message-passing design overlaps computation and communication, reducing end-to-end verification complexity to \(O(n/P + \log P)\). Consequently, PROSIMA achieves sub-linear consensus latency compared with conventional full-ledger replication while preserving Byzantine fault tolerance and verifiable trust.



\section{Evaluation}
\subsection{Experimental Setup}
\subsubsection{Testbed} 

We evaluated \textsc{Prosima} on a testbed with 20 virtualized compute nodes, each provisioned with 8 CPU threads (Intel Xeon 3.0~GHz) and 32~GB of DDR4 memory. All nodes share a centralized {NVIDIA RTX~A6000 GPU} with 48~GB of VRAM, enabling concurrent execution of federated training, semantic reconstruction, and on-chain verification. This configuration approximates a realistic distributed edge environment while maintaining consistent compute and memory budgets per node. Each node runs an identical software stack based on Ubuntu~22.04, Python~3.10, PyTorch~2.2, and CUDA~12.1 to ensure deterministic reproducibility. The blockchain framework is integrated with the {MPI}-based parallel programming model~\cite{mpi-online} to facilitate low-latency inter-node communication. To further enhance runtime efficiency, we employ an edge-aware caching mechanism for latent fingerprints and semantic embeddings, storing frequently accessed vectors in GPU-adjacent high-bandwidth memory to minimize PCIe transfer overhead and maximize data locality during shard retrieval and consensus operations.



\begin{figure}[!t]
\centering
    \centering
    \includegraphics[width={0.69\columnwidth}]
    {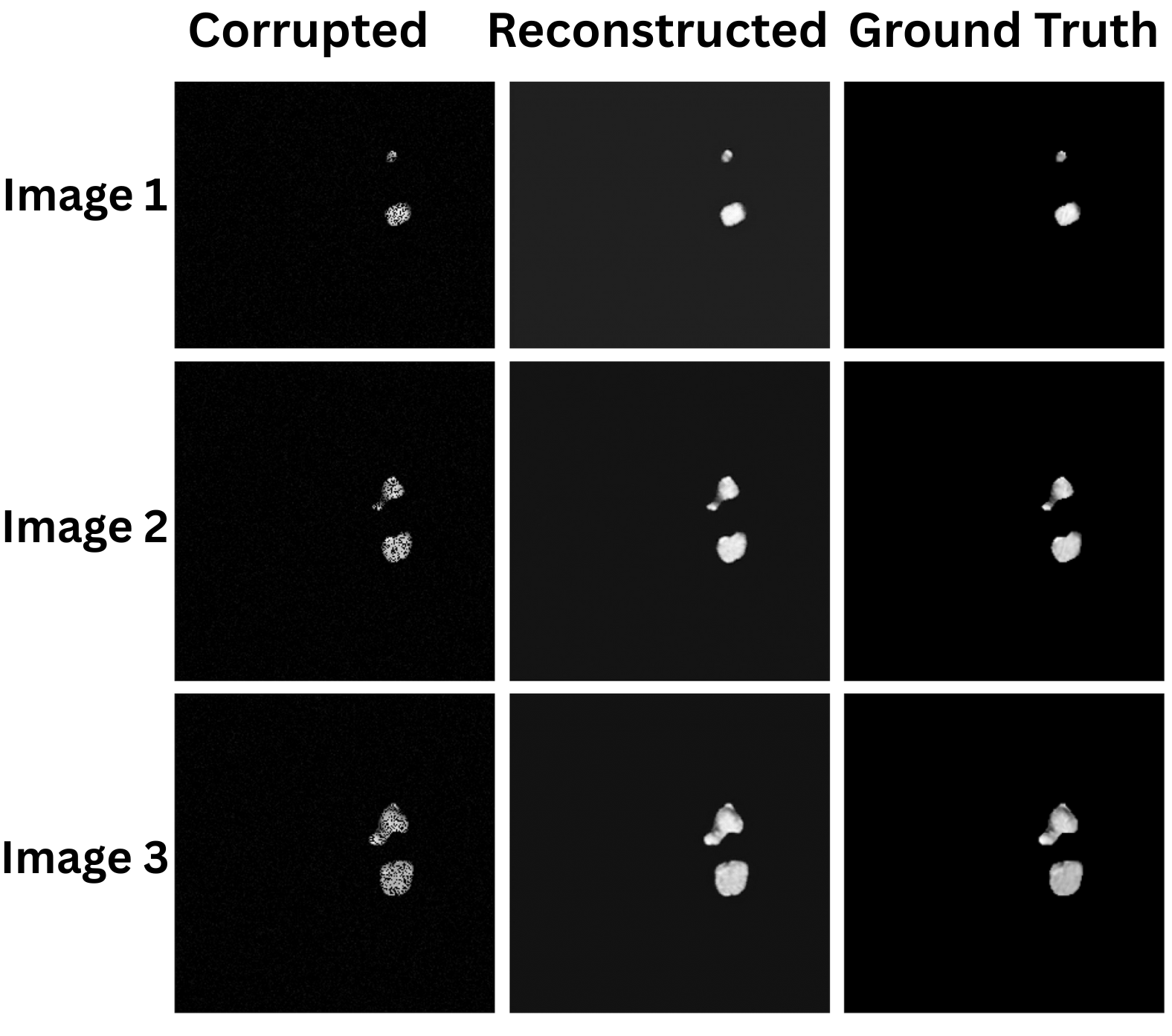}
    \caption{Visual
fidelity achieved by the proposed \textsc{Prosima} framework on three different representative corrupted brain tumor images.}
    \label{fig:qualitative-image}
\vspace{-5mm}
\end{figure}


\subsubsection{Dataset}
We evaluate our semantic-aware reconstruction framework on the BraTS 2020 (Brain Tumor Segmentation) challenge dataset~\cite{menze2015brats,bakas2017advancing,bakas2018identifying}. BraTS provides multi-institutional, pre-operative, multimodal brain MRI scans with expert tumor annotations, including T1, contrast-enhanced T1 (T1c), T2, and FLAIR sequences.

In our experiments, we extract 2D axial slices of size $256 \times 256$ from the BraTS 2020 training set. We use 1{,}000 slices for training and evaluation, covering a diverse range of tumor locations, sizes, and surrounding healthy tissue. This setting stresses the ability of our reconstruction model to preserve fine-grained anatomical structures (e.g., tumor core, edema, and normal brain parenchyma) under corruption, making BraTS an appropriate benchmark for semantic-consistent medical image reconstruction.

\subsubsection{Evaluated Systems}
We compare \textsc{Prosima} with the multiple models and variants to measure reconstruction quality and trust: 
\begin{itemize}
    \item[\textbf{A.}] \textit{\textbf{U-Net (pixel)}}: A standard U-Net trained with L1/L2 pixel-wise reconstruction loss.  
    \item[\textbf{B.}] \textit{\textbf{U-Net + perceptual}}: The same U-Net architecture trained with an additional VGG-based perceptual loss to enhance structural fidelity.  
    \item[\textbf{C.}] \textit{\textbf{U-Net + semantic}}: A semantic-aware U-Net conditioned on high-level semantic embeddings using the proposed semantic loss, but without latent fingerprint conditioning (ablation).  
    \item[\textbf{D.}] \textit{\textbf{PROSIMA}}: The proposed Hybrid U-Net with latent fingerprint conditioning, semantic loss, GFT-based sharding, and blockchain-based provenance anchoring.  
\end{itemize}

\noindent Additionally, we include an ablation variant, \textit{PROSIMA (w/o GFT)}, which retains the full semantic-hybrid model but replaces GFT-based sharding with random duplication for system-level comparison.



\subsection{Results and Discussion}

\subsubsection{Qualitative Evaluation}


Fig.~\ref{fig:qualitative-image} illustrates the visual fidelity achieved by the proposed \textsc{Prosima} framework on representative corrupted samples. 
The reconstructed outputs accurately recover structural integrity and preserve fine-scale tissue boundaries even under severe signal degradation. 
Compared to pixel-only baselines, \textsc{Prosima} produces noticeably smoother intensity transitions, consistent lesion morphologies, and preserved anatomical contours.
This superior reconstruction quality stems from the semantic-aware conditioning introduced during federated training and the deterministic fingerprint-guided retrieval that enforces latent consistency.
As a result, the recovered slices exhibit both pixel-level realism and semantic alignment with the ground truth, confirming the model’s ability to integrate low-level texture recovery with high-level contextual reasoning.

\subsubsection{Quantitative Evaluation}

Table~\ref{tab:main_results} presents the overall reconstruction and provenance performance across all evaluated models. 
The proposed \textsc{Prosima} framework consistently achieves superior quantitative results, obtaining the highest PSNR ($\ge 28.2$~dB), SSIM ($\ge 0.86$), and semantic embedding cosine ($\ge 0.92$) while maintaining a $100\%$ verification rate with the lowest latency of approximately $204$~ms per image. 
The progressive improvements from pixel-based to semantic-guided U-Net variants confirm the effectiveness of incorporating perceptual and semantic supervision for structurally consistent reconstructions. 
Compared with its w/o-GFT counterpart, the full \textsc{Prosima} configuration further reduces latency through geometric fragmentation, which limits redundant replication and accelerates distributed shard retrieval. 
These results collectively demonstrate that \textsc{Prosima} achieves a strong balance between high-fidelity image reconstruction, semantic alignment, and blockchain-based verifiability, delivering dependable and scalable medical image recovery beyond traditional deep reconstruction baselines.

\begin{table}[!t]
\centering
\scriptsize
 \caption{Embedding Cosine vs Gaussian noise severity (mean).} 
\label{tab:cosine_vs_noise}
\begin{tabular}{lccc}

\toprule
Method / Noise $\sigma$ & 0.02 & 0.05 & 0.10 \\
\midrule
U\text{-}Net (pixel)        & $\ge 0.56$ & $\ge 0.45$ & $\ge 0.39$ \\
U\text{-}Net + perceptual   & $\ge 0.63$ & $\ge 0.52$ & $\ge 0.42$ \\
U\text{-}Net + semantic     & $\ge 0.86$ & $\ge 0.81$ & $\ge 0.73$ \\
PROSIMA (full)              & $\ge 0.91$ & $\ge 0.86$ & $\ge 0.79$ \\
\bottomrule

\end{tabular}
\vspace{-2mm}
\end{table}

\begin{table}[t]
\centering
\scriptsize
\caption{Ablation results (Embedding Cosine change and latency). Values are mean over test set. Removing semantic loss drops cosine dramatically; removing fingerprint only affects verifiability (and slightly semantic alignment); replacing GFT increases latency markedly.}
\label{tab:ablation}
\begin{tabular}{lcc}
\toprule
Variant & Embed. Cosine & Latency (ms) \\
\midrule
\textsc{Prosima} (full)                 & $\ge 0.91$ & $\le 201$ \\
w/o semantic loss ($\lambda_2{=}0$)   & $\ge 0.63$ & $\le 200$ \\
w/o fingerprint     & $\ge 0.86$ & $\le 181$ \\
w/o GFT (random dup)    & $\ge 0.88$ & $\le 263$ \\
\bottomrule
\end{tabular}
\end{table}




\begin{table}[!t]
\centering
\scriptsize
\caption{Systems-level comparison (N=20 nodes).}
\label{tab:systems}
\begin{tabular}{lccc}

\toprule
System & storage/node (MB) & Repl. factor & latency (ms) \\
\midrule
Full-ledger (naïve)      & $0.21$ & $20.43$ & $650.0 \pm 0.0$ \\
Random-shard + dup       & $0.04$ & $3.59$  & $257.7 \pm 5.3$ \\
\textbf{\textsc{Prosima} (GFT)}   & $\mathbf{0.02}$ & $\mathbf{1.28}$ & $\mathbf{227.8 \pm 3.3}$ \\
\bottomrule

\end{tabular}
\vspace{-5mm}
\end{table}



\subsubsection{Robustness Under Real-World Corruptions}
To test robustness, we apply additive Gaussian noise ($\mu=0$, $\sigma=0.05$) as a representative real-world corruption. Gaussian noise reflects natural sensor and environmental perturbations and has been widely used in robustness studies \cite{gaussian-noise-1,gaussian-noise-2}. In contrast, random masking, which removes or replaces large portions of pixels, is less common in realistic scenarios and thus was not considered in this work. We also do not include adversarial perturbations (e.g., FGSM-Fast Gradient Sign Method~\cite{fgsm}) because our focus is on realistic, naturally occurring corruptions rather than worst-case attacks, which are typically out of the scope of normal operational scenarios~\cite{gaussian-noise-1,gaussian-noise-2}.

Table~\ref{tab:cosine_vs_noise} evaluates the robustness of different reconstruction models under varying levels of Gaussian noise, reflecting real-world sensor degradation and acquisition imperfections. 
While pixel-based U-Net baselines show a rapid decline in embedding cosine similarity as noise severity increases ($0.56 \rightarrow 0.39$), semantic-guided variants exhibit significantly higher resilience. 
The proposed \textsc{Prosima} framework maintains stable semantic alignment even under strong corruptions, achieving an embedding cosine of $\ge 0.79$ at $\sigma=0.10$, compared to $\le 0.45$ for purely pixel-based methods. 
This robustness arises from \textsc{Prosima}’s latent fingerprint anchoring and semantic conditioning, which constrain the reconstruction manifold toward contextually meaningful representations rather than pixel-level memorization. 
These results demonstrate that \textsc{Prosima} not only restores visual fidelity but also preserves the underlying semantic integrity of medical images under challenging, noise-heavy conditions, ensuring reliable reconstruction performance in real clinical scenarios.

\subsubsection{Ablation Studies}

Table~\ref{tab:ablation} highlights the contribution of each major component in \textsc{Prosima}. 
Removing the semantic loss sharply decreases embedding alignment (from $0.91$ to $0.63$), confirming the necessity of semantic supervision for perceptually coherent reconstructions. 
Disabling fingerprint conditioning slightly reduces embedding consistency but compromises verifiability, while omitting GFT leads to higher latency due to inefficient shard placement. 
Overall, the complete PROSIMA configuration achieves the best balance between reconstruction quality, provenance integrity, and system efficiency, validating its design as a robust, trustworthy, and scalable medical image reconstruction framework.

\subsubsection{Blockchain Memory Efficiency}
Table~\ref{tab:systems} compares the storage footprint and replication efficiency of \textsc{Prosima} against traditional blockchain baselines across a 20-node network. 
The proposed GFT-based design achieves the lowest storage overhead, requiring only $0.02$~MB per node and a replication factor of $1.28\times$, compared to $20.43\times$ in naive full-ledger replication. 
This substantial reduction arises from \textsc{Prosima}’s geometric fragmentation and selective shard anchoring, which distribute only essential provenance metadata instead of duplicating the full image payload. 
Despite the significant memory savings, the system maintains a low average commit latency of $228$~ms, demonstrating that storage efficiency does not compromise transaction throughput. 
These results confirm that \textsc{Prosima}’s ledger layer offers a scalable and memory-efficient blockchain backbone, capable of supporting high-volume medical image verification without incurring the prohibitive redundancy typical of conventional distributed ledgers.


\begin{figure}[!t]
\centering
\begin{subfigure}[b]{0.82\columnwidth}
    \centering
    \includegraphics[width=\textwidth]{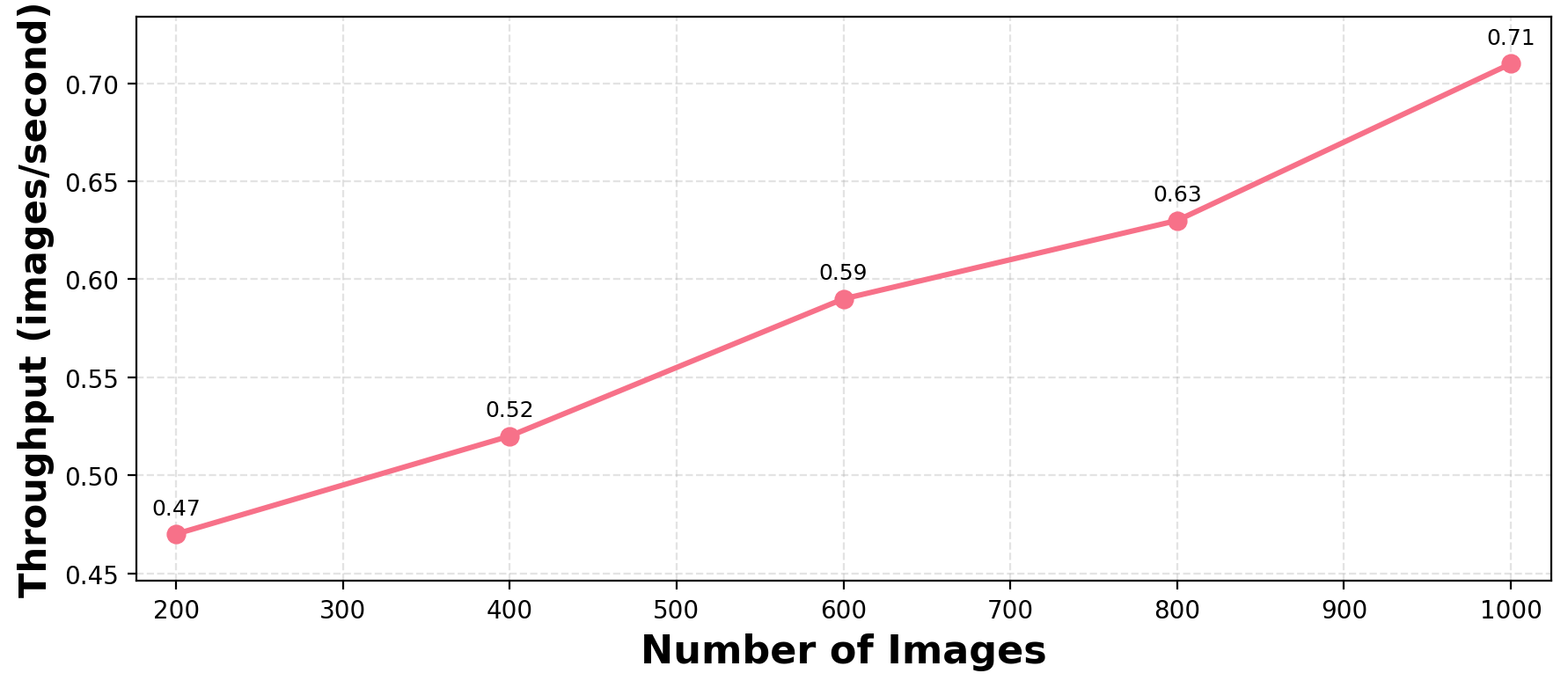}
    \caption{Throughput scalability.}
    \label{fig:throughput-scalability}
\end{subfigure}
\hfill
\begin{subfigure}[b]{0.82\columnwidth}
    \centering
    \includegraphics[width=\textwidth]{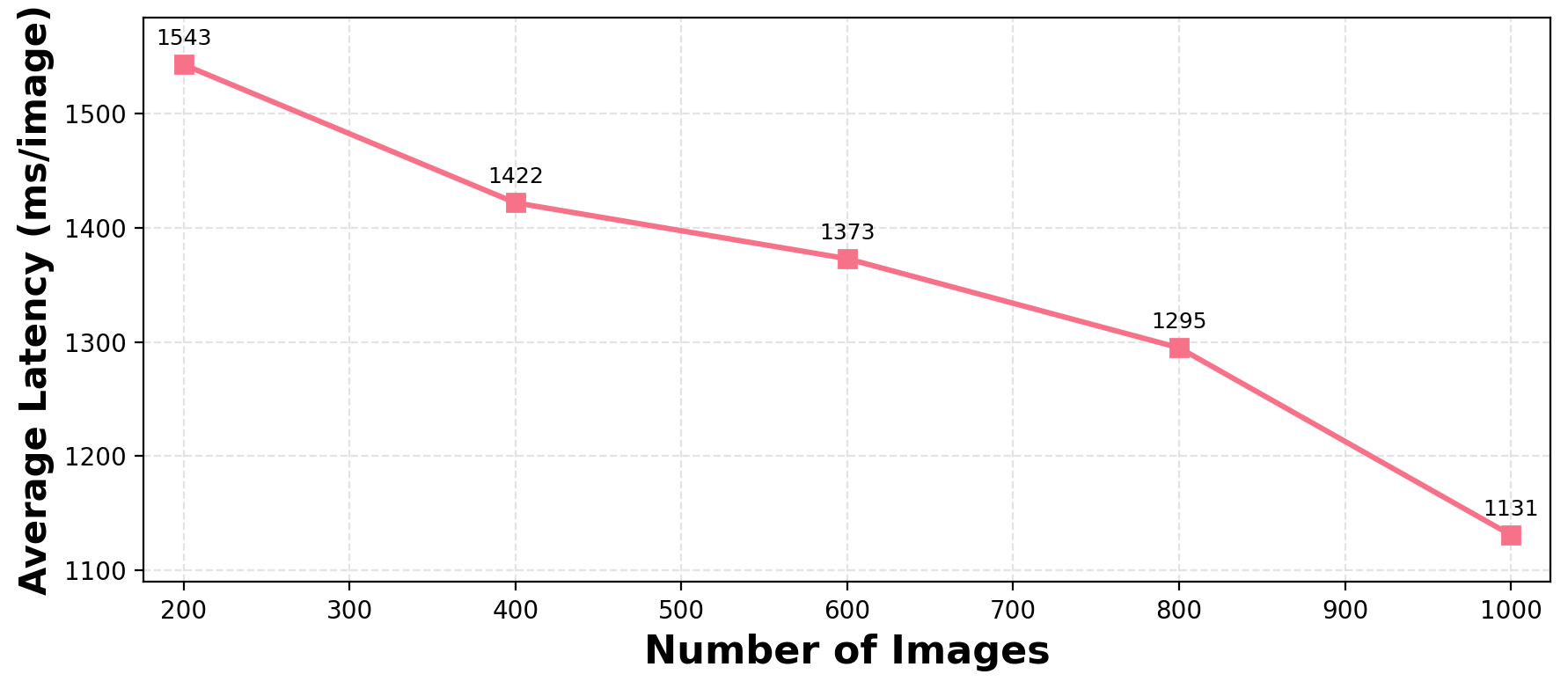}
    \caption{Latency scalability.}
    \label{fig:latency-scalability}
\end{subfigure}
\vspace{-2mm}
\caption{Blockchain throughput and latency scalability of \textsc{Prosima}.}
\label{fig:scalability}
\vspace{-7mm}
\end{figure}

\vspace{-2mm}
\subsubsection{Blockchain Throughput and Latency Scalability}
\vspace{-2mm}
We further evaluate the scalability of the \textsc{Prosima} ledger by analyzing its throughput and per-image latency across increasing workloads ranging from 200 to 1000 images. 
As illustrated in Fig.~\ref{fig:scalability}, throughput rises consistently from 0.47 to 0.71~images/s, while the average latency per image decreases from 1543~ms to 1131~ms. 
This complementary trend reflects the efficiency of \textsc{Prosima}’s MPI-parallel verification pipeline and scale-free sharding strategy, which jointly minimize synchronization bottlenecks and exploit locality-aware workload distribution. 
Unlike traditional blockchain frameworks that suffer linear degradation under load, \textsc{Prosima} demonstrates sub-linear latency scaling and stable throughput growth, confirming that its hybrid geometric fragmentation and consensus design supports high-throughput, low-latency provenance verification at scale.


\vspace{-4mm}
\section{Conclusion and Future Work}
\vspace{-2mm}
This work presented \textsc{Prosima}, a provenance-aware semantic reconstruction framework that unifies deep generative modeling with blockchain-based trust verification. 
Through extensive experiments, we demonstrated that semantic conditioning significantly enhances structural and perceptual fidelity under various corruption levels, while latent fingerprint anchoring ensures verifiable and tamper-evident provenance. 
Our system achieves high robustness and low-latency consensus through an MPI-accelerated, scale-free ledger architecture, bridging the gap between efficient image recovery and trustworthy distributed AI. 
Future extensions will explore tighter coupling with decentralized storage networks and adaptive on-device verification within heterogeneous edge environments, further advancing reliable, privacy-preserving medical image reconstruction at scale.

{\small
\bibliographystyle{ieeenat_fullname}
\bibliography{ref}

@article{azad2024medical,
  title={Medical image segmentation review: The success of u-net},
  author={Azad, Reza and Aghdam, Ehsan Khodapanah and Rauland, Amelie and Jia, Yiwei and Avval, Atlas Haddadi and Bozorgpour, Afshin and Karimijafarbigloo, Sanaz and Cohen, Joseph Paul and Adeli, Ehsan and Merhof, Dorit},
  journal={IEEE Transactions on Pattern Analysis and Machine Intelligence},
  year={2024},
  publisher={IEEE}
}

@article{du2020medical,
  title={Medical image segmentation based on U-net: A review.},
  author={Du, Getao and Cao, Xu and Liang, Jimin and Chen, Xueli and Zhan, Yonghua},
  journal={Journal of Imaging Science \& Technology},
  volume={64},
  number={2},
  year={2020}
}

@misc{mpi-online,
  title = {{MPI} Forum Documentation},
  author = {{Message Passing Interface Forum}},
  howpublished = {},
  note = {Accessed: 2025-11-13}
}

@article{fgsm,
  title={Trans-IFFT-FGSM: a novel fast gradient sign method for adversarial attacks},
  author={Naseem, Muhammad Luqman},
  journal={Multimedia Tools and Applications},
  volume={83},
  number={29},
  pages={72279--72299},
  year={2024},
  publisher={Springer}
}

@inproceedings{gaussian-noise-1,
  title={Improving robustness against common corruptions with frequency biased models},
  author={Saikia, Tonmoy and Schmid, Cordelia and Brox, Thomas},
  booktitle={Proceedings of the IEEE/CVF International Conference on Computer Vision},
  pages={10211--10220},
  year={2021}
}

@inproceedings{gaussian-noise-2,
  title={Adversarial examples are a natural consequence of test error in noise},
  author={Gilmer, Justin and Ford, Nicolas and Carlini, Nicholas and Cubuk, Ekin},
  booktitle={International Conference on Machine Learning},
  pages={2280--2289},
  year={2019},
  organization={PMLR}
}

@article{song2021solving,
  title={Solving inverse problems in medical imaging with score-based generative models},
  author={Song, Yang and Shen, Liyue and Xing, Lei and Ermon, Stefano},
  journal={arXiv preprint arXiv:2111.08005},
  year={2021}
}

@inproceedings{ronneberger2015u,
  title={U-net: Convolutional networks for biomedical image segmentation},
  author={Ronneberger, Olaf and Fischer, Philipp and Brox, Thomas},
  booktitle={International Conference on Medical image computing and computer-assisted intervention},
  pages={234--241},
  year={2015},
  organization={Springer}
}

@article{kingma2014adam,
  title={Adam: A method for stochastic optimization},
  author={Kingma, Diederik P},
  journal={arXiv preprint arXiv:1412.6980},
  year={2014}
}

@inproceedings{attack-1,
  title={$\{$CT-GAN$\}$: Malicious tampering of 3d medical imagery using deep learning},
  author={Mirsky, Yisroel and Mahler, Tom and Shelef, Ilan and Elovici, Yuval},
  booktitle={28th USENIX Security Symposium (USENIX Security 19)},
  pages={461--478},
  year={2019}
}

@article{attack-2,
  title={Medical images under tampering},
  author={Tsai, Min-Jen and Lin, Ping-Ying},
  journal={Multimedia Tools and Applications},
  volume={83},
  number={24},
  pages={65407--65439},
  year={2024},
  publisher={Springer}
}

@inproceedings{gan-attack-detect-1,
  title={Detection of GAN-manipulated medical images through deep learning techniques},
  author={Aruna, S and Narayan, Surabhi},
  booktitle={2024 International Conference on Advances in Modern Age Technologies for Health and Engineering Science (AMATHE)},
  pages={1--6},
  year={2024},
  organization={IEEE}
}

@article{gan-attack-detect-2,
  title={GAN-based medical image small region forgery detection via a two-stage cascade framework},
  author={Zhang, Jianyi and Huang, Xuanxi and Liu, Yaqi and Han, Yuyang and Xiang, Zixiao},
  journal={Plos one},
  volume={19},
  number={1},
  pages={e0290303},
  year={2024},
  publisher={Public Library of Science San Francisco, CA USA}
}

@article{gan-attack-detect-3,
  title={MITS-GAN: Safeguarding medical imaging from tampering with generative adversarial networks},
  author={Pasqualino, Giovanni and Guarnera, Luca and Ortis, Alessandro and Battiato, Sebastiano},
  journal={Computers in Biology and Medicine},
  volume={183},
  pages={109248},
  year={2024},
  publisher={Elsevier}
}

@article{attack-detect-4,
  title={Security and Deep Learning: Verifying the Authenticity of CT Images},
  author={Gerlofsma, Markus and Petkovic, Milan and van Liesdonk, Ir Peter},
  year={2019}
}

@article{barabasi2003scale,
  title={Scale-free networks},
  author={Barab{\'a}si, Albert-L{\'a}szl{\'o} and Bonabeau, Eric},
  journal={Scientific american},
  volume={288},
  number={5},
  pages={60--69},
  year={2003},
  publisher={JSTOR}
}

@article{abdella2021architecture,
  title={An architecture and performance evaluation of blockchain-based peer-to-peer energy trading},
  author={Abdella, Juhar and Tari, Zahir and Anwar, Adnan and Mahmood, Abdun and Han, Fengling},
  journal={IEEE Transactions on Smart Grid},
  volume={12},
  number={4},
  pages={3364--3378},
  year={2021},
  publisher={IEEE}
}

@article{alskaif2021blockchain,
  title={Blockchain-based fully peer-to-peer energy trading strategies for residential energy systems},
  author={AlSkaif, Tarek and Crespo-Vazquez, Jose L and Sekuloski, Milos and Van Leeuwen, Gijs and Catal{\~a}o, Jo{\~a}o PS},
  journal={IEEE Transactions on Industrial Informatics},
  volume={18},
  number={1},
  pages={231--241},
  year={2021},
  publisher={IEEE}
}

@article{nguyen2022latency,
  title={Latency optimization for blockchain-empowered federated learning in multi-server edge computing},
  author={Nguyen, Dinh C and Hosseinalipour, Seyyedali and Love, David J and Pathirana, Pubudu N and Brinton, Christopher G},
  journal={IEEE Journal on Selected Areas in Communications},
  volume={40},
  number={12},
  pages={3373--3390},
  year={2022},
  publisher={IEEE}
}

@article{wang2023lightweight,
  title={Lightweight and secure data sharing based on proxy re-encryption for blockchain-enabled industrial internet of things},
  author={Wang, Fengqun and Cui, Jie and Zhang, Qingyang and He, Debiao and Gu, Chengjie and Zhong, Hong},
  journal={IEEE Internet of Things Journal},
  volume={11},
  number={8},
  pages={14115--14126},
  year={2023},
  publisher={IEEE}
}

@inproceedings{mamun2021baash,
  title={BAASH: Lightweight, efficient, and reliable blockchain-as-a-service for hpc systems},
  author={Mamun, Abdullah Al and Yan, Feng and Zhao, Dongfang},
  booktitle={Proceedings of the International Conference for High Performance Computing, Networking, Storage and Analysis},
  pages={1--18},
  year={2021}
}

@inproceedings{al2022dean,
  title={Dean: A lightweight and resource-efficient blockchain protocol for reliable edge computing},
  author={Al-Mamun, Abdullah and Shen, Haoting and Zhao, Dongfang},
  booktitle={2022 IEEE International Parallel and Distributed Processing Symposium (IPDPS)},
  pages={1261--1271},
  year={2022},
  organization={IEEE}
}

@inproceedings{zhang2020cycledger,
  author = {Zhang, Mengqian and Li, Jichen and Chen, Zhaohua and Chen, Hongyin and Deng, Xiaotie},
  title = {CycLedger: A Scalable and Secure Parallel Protocol for Distributed Ledger via Sharding},
  booktitle = {Proceedings of the IEEE International Parallel and Distributed Processing Symposium},
  series = {IPDPS '20},
  year = {2020},
  pages = {358--367},
  publisher = {IEEE},
  doi = {10.1109/IPDPS47924.2020.00046}
}

@inproceedings{kokoris2018omniledger,
  author = {Kokoris-Kogias, Eleftherios and Jovanovic, Philipp and Gasser, Linus and Gailly, Nicolas and Syta, Ewa and Ford, Bryan},
  title = {OmniLedger: A Secure, Scale-Out, Decentralized Ledger via Sharding},
  booktitle = {IEEE Symposium on Security and Privacy},
  series = {SP '18},
  year = {2018},
  pages = {583--598},
  publisher = {IEEE},
  doi = {10.1109/SP.2018.000-5}
}

@inproceedings{defussion,
  title={Diffusion-based adaptation for classification of unknown degraded images},
  author={Daultani, Dinesh and Tanaka, Masayuki and Okutomi, Masatoshi and Endo, Kazuki},
  booktitle={Proceedings of the IEEE/CVF Conference on Computer Vision and Pattern Recognition},
  pages={5982--5991},
  year={2024}
}

@article{gan-2,
  title={MSC-GAN: A Multi-Stream Complementary Generative Adversarial Network with Grouping Learning for Multitemporal Cloud Removal},
  author={Zhou, Haoran and Wang, Yanjiang and Liu, Weifeng and Tao, Dapeng and Ma, Wei and Liu, Baodi},
  journal={IEEE Transactions on Geoscience and Remote Sensing},
  year={2024},
  publisher={IEEE}
}

@article{gan-1,
  title={Reconstruction of large-scale missing data in remote sensing images using Extend-GAN},
  author={Cui, Yongchuan and Liu, Peng and Song, Bingze and Zhao, Lingjun and Ma, Yan and Chen, Lajiao},
  journal={IEEE Geoscience and Remote Sensing Letters},
  volume={21},
  pages={1--5},
  year={2024},
  publisher={IEEE}
}

@article{unet-1,
  title={Load image inpainting: An improved U-Net based load missing data recovery method},
  author={Liu, Liqi and Liu, Yanli},
  journal={Applied Energy},
  volume={327},
  pages={119988},
  year={2022},
  publisher={Elsevier}
}

@article{unet-2,
  title={MICU: Image super-resolution via multi-level information compensation and U-net},
  author={Chen, Yuantao and Xia, Runlong and Yang, Kai and Zou, Ke},
  journal={Expert Systems with Applications},
  volume={245},
  pages={123111},
  year={2024},
  publisher={Elsevier}
}

@article{dl-image-recovery,
  title={Deep learning with noisy labels: Exploring techniques and remedies in medical image analysis},
  author={Karimi, Davood and Dou, Haoran and Warfield, Simon K and Gholipour, Ali},
  journal={Medical image analysis},
  volume={65},
  pages={101759},
  year={2020},
  publisher={Elsevier}
}

@article{compression-1,
  title={Deep learning-assisted medical image compression challenges and opportunities: systematic review},
  author={Bourai, Nour El Houda and Merouani, Hayet Farida and Djebbar, Akila},
  journal={Neural Computing and Applications},
  volume={36},
  number={17},
  pages={10067--10108},
  year={2024},
  publisher={Springer}
}

@article{noise,
  title={Noise issues prevailing in various types of medical images},
  author={Goyal, Bhawna and Agrawal, Sunil and Sohi, BS},
  journal={Biomedical \& Pharmacology Journal},
  volume={11},
  number={3},
  pages={1227},
  year={2018},
  publisher={Biomedical and Pharmacology Journal}
}

@inproceedings{tampering-3,
  title={Medical Image Tampering Detection using Deep Learning},
  author={Reddy, B Ramasubba and Kumar, M Sunil and Neelima, P and Sushama, C and Sailaja, Vedala Naga and Ganesh, D},
  booktitle={2024 5th International Conference on Smart Electronics and Communication (ICOSEC)},
  pages={1480--1485},
  year={2024},
  organization={IEEE}
}

@article{tampering-2,
  title={Medical image tamper detection based on passive image authentication},
  author={Ulutas, Guzin and Ustubioglu, Arda and Ustubioglu, Beste and V. Nabiyev, Vasif and Ulutas, Mustafa},
  journal={Journal of digital imaging},
  volume={30},
  number={6},
  pages={695--709},
  year={2017},
  publisher={Springer}
}

@article{tampering-1,
  title={Medical image watermarking for ownership \& tamper detection},
  author={Alshanbari, Hanan S},
  journal={Multimedia tools and applications},
  volume={80},
  number={11},
  pages={16549--16564},
  year={2021},
  publisher={Springer}
}

@article{yaqoob2019security,
  title={Security vulnerabilities, attacks, countermeasures, and regulations of networked medical devices—A review},
  author={Yaqoob, Tahreem and Abbas, Haider and Atiquzzaman, Mohammed},
  journal={IEEE Communications Surveys \& Tutorials},
  volume={21},
  number={4},
  pages={3723--3768},
  year={2019},
  publisher={IEEE}
}

@article{hussain2022modern,
  title={Modern diagnostic imaging technique applications and risk factors in the medical field: a review},
  author={Hussain, Shah and Mubeen, Iqra and Ullah, Niamat and Shah, Syed Shahab Ud Din and Khan, Bakhtawar Abduljalil and Zahoor, Muhammad and Ullah, Riaz and Khan, Farhat Ali and Sultan, Mujeeb A},
  journal={BioMed research international},
  volume={2022},
  number={1},
  pages={5164970},
  year={2022},
  publisher={Wiley Online Library}
}

@inproceedings{jin2017fbpconvnet,
  title={Deep Convolutional Neural Network for Inverse Problems in Imaging},
  author={Jin, Kyong Hwan and McCann, Michael T. and Froustey, Emmanuel and Unser, Michael},
  booktitle={IEEE Transactions on Image Processing},
  year={2017},
  volume={26},
  number={9},
  pages={4509--4522},
  doi={10.1109/TIP.2017.2713099}
}

@inproceedings{yang2016admmnet,
  title={{ADMM-Net}: A Deep Learning Approach for Compressive Sensing MRI},
  author={Yang, Yan and Sun, Jian and Li, Huibin and Xu, Zongben},
  booktitle={Advances in Neural Information Processing Systems (NeurIPS)},
  year={2016}
}

@article{hammernik2018varnet,
  title={Learning a Variational Network for Reconstruction of Accelerated MRI Data},
  author={Hammernik, Kerstin and Klatzer, Teresa and Knoll, Florian and Sodickson, Daniel K. and Pock, Thomas and Maier, Andreas},
  journal={Magnetic Resonance in Medicine},
  year={2018},
  volume={79},
  number={6},
  pages={3055--3071},
  doi={10.1002/mrm.26977}
}

@inproceedings{adler2018lpd,
  title={Learned Primal-Dual Reconstruction},
  author={Adler, Jonas and {\"O}ktem, Ozan},
  booktitle={IEEE/CVF Conference on Computer Vision and Pattern Recognition (CVPR) Workshops},
  year={2018}
}

@article{mccann2017review,
  title={Convolutional Neural Networks for Inverse Problems in Imaging: A Review},
  author={McCann, Michael T. and Jin, Kyong Hwan and Unser, Michael},
  journal={IEEE Signal Processing Magazine},
  volume={34},
  number={6},
  pages={85--95},
  year={2017},
  doi={10.1109/MSP.2017.2739299}
}

@inproceedings{venkatakrishnan2013pnp,
  title={Plug-and-Play Priors for Model Based Reconstruction},
  author={Venkatakrishnan, Sreehari and Bouman, Charles A. and Wohlberg, Brendt},
  booktitle={IEEE Global Conference on Signal and Information Processing},
  pages={945--948},
  year={2013},
  doi={10.1109/GlobalSIP.2013.6737048}
}

@article{chan2016pnp,
  title={Plug-and-Play {ADMM} for Image Restoration: Fixed-Point Convergence and Applications},
  author={Chan, Stanley H. and Wang, Xiran and Elgendy, Omar A.},
  journal={IEEE Transactions on Computational Imaging},
  volume={3},
  number={1},
  pages={84--98},
  year={2017},
  doi={10.1109/TCI.2016.2626538}
}

@article{romano2017red,
  title={The Little Engine That Could: Regularization by Denoising ({RED})},
  author={Romano, Yaniv and Elad, Michael and Milanfar, Peyman},
  journal={SIAM Journal on Imaging Sciences},
  volume={10},
  number={4},
  pages={1804--1844},
  year={2017},
  doi={10.1137/16M1102884}
}

@inproceedings{song2021scoremed,
  title={Solving Inverse Problems in Medical Imaging with Score-Based Generative Models},
  author={Song, Yang and Shen, Liyue and Xing, Lei and Ermon, Stefano},
  booktitle={International Conference on Learning Representations (ICLR)},
  year={2022}
}

@inproceedings{chung2023parallel,
  title={Parallel Diffusion Models of Operator and Image for Blind Inverse Problems},
  author={Chung, Hyungjin and Sim, Byeongsu and Ye, Jong Chul},
  booktitle={IEEE/CVF Conference on Computer Vision and Pattern Recognition (CVPR)},
  year={2023},
  pages={3398--3408},
  doi={10.1109/CVPR52729.2023.00330}
}

@misc{dds_github,
  title={{DDS}: Official PyTorch Implementation},
  author={Chung, Hyungjin},
  howpublished={GitHub repository},
  year={2023}
}

@inproceedings{boecking2022biovil,
  title={Making the Most of Text Semantics to Improve Biomedical Vision--Language Processing},
  author={Boecking, Benedikt and Usuyama, Naoto and Bannur, Shruthi and Castro, Daniel C. and Schwaighofer, Anton and Hyland, Stephanie and Wetscherek, Maria and Naumann, Tristan and Nori, Aditya and Alvarez-Valle, Javier and Poon, Hoifung and Oktay, Ozan},
  booktitle={European Conference on Computer Vision (ECCV)},
  year={2022}
}

@inproceedings{bannur2023biovilt,
  title={Learning to Exploit Temporal Structure for Biomedical Vision--Language Processing},
  author={Bannur, Shruthi and Boecking, Benedikt and Usuyama, Naoto and others},
  booktitle={IEEE/CVF Conference on Computer Vision and Pattern Recognition (CVPR)},
  year={2023}
}

@inproceedings{medclip_emnlp,
  title={{MedCLIP}: Contrastive Learning from Unpaired Medical Images and Text},
  author={Wang, Jiaqi and Liu, Xiaosong and Lin, Xin and others},
  booktitle={EMNLP},
  year={2022}
}

@article{medclip_pmc,
  title={{MedCLIP}: Contrastive Learning from Unpaired Medical Images and Text},
  author={Wang, Jiaqi and Liu, Xiaosong and Lin, Xin and others},
  journal={Patterns},
  year={2023}
}

@article{hasani2022trustworthy,
  title={Trustworthy artificial intelligence in medical imaging},
  author={Hasani, Navid and Morris, Michael A and Rhamim, Arman and Summers, Ronald M and Jones, Elizabeth and Siegel, Eliot and Saboury, Babak},
  journal={PET clinics},
  volume={17},
  number={1},
  pages={1},
  year={2022}
}

@article{esr_blockchain_whitepaper,
  title={Blockchain and Medical Imaging: A White Paper of the European Society of Radiology},
  author={{European Society of Radiology (ESR)}},
  journal={Insights into Imaging},
  volume={12},
  number={1},
  pages={124},
  year={2021},
  doi={10.1186/s13244-021-01029-y}
}

@inproceedings{castro1999pbft,
  title={Practical Byzantine Fault Tolerance},
  author={Castro, Miguel and Liskov, Barbara},
  booktitle={OSDI},
  year={1999},
  pages={173--186}
}

@phdthesis{buchman2016tendermint,
  title={Tendermint: Byzantine Fault Tolerance in the Age of Blockchains},
  author={Buchman, Ethan},
  school={University of Guelph},
  year={2016}
}

@inproceedings{yin2019hotstuff,
  title={HotStuff: BFT Consensus in the Lens of Blockchain},
  author={Yin, Maofan and Malkhi, Dahlia and Reiter, Michael K. and Gueta, Guy Golan and Abraham, Ittai},
  booktitle={ACM PODC},
  year={2019},
  pages={347--356},
  doi={10.1145/3293611.3331591}
}

@article{menze2015brats,
  title        = {The Multimodal Brain Tumor Image Segmentation Benchmark (BRATS)},
  author       = {Menze, Bjoern H and Jakab, Andras and Bauer, Stefan and Kalpathy-Cramer, Jayashree and Farahani, Keyvan and Kirby, Justin and et al.},
  journal      = {IEEE Transactions on Medical Imaging},
  volume       = {34},
  number       = {10},
  pages        = {1993--2024},
  year         = {2015},
  doi          = {10.1109/TMI.2014.2377694}
}

@article{bakas2017advancing,
  title        = {Advancing The Cancer Genome Atlas glioma MRI collections with expert segmentation labels and radiomic features},
  author       = {Bakas, Spyridon and Akbari, Hamed and Sotiras, Aristeidis and Bilello, Michel and Rozycki, Michal and Kirby, Justin and et al.},
  journal      = {Scientific Data},
  volume       = {4},
  pages        = {170117},
  year         = {2017},
  doi          = {10.1038/sdata.2017.117}
}

@article{bakas2018identifying,
  title        = {Identifying the Best Machine Learning Algorithms for Brain Tumor Segmentation, Progression Assessment, and Overall Survival Prediction in the BRATS Challenge},
  author       = {Bakas, Spyridon and Reyes, Mauricio and Jakab, Andras and Bauer, Stefan and Rempfler, Markus and Crimi, Alessandro and et al.},
  journal      = {arXiv preprint},
  year         = {2018},
  eprint       = {1811.02629},
  archivePrefix= {arXiv}
}
}

\end{document}